# Validation of a Facial-Video Digital Hydration Biomarker Score for Tracking Hydration State Transitions in a Real-World Fasting Model

Yudara Kularathne, D.M.J. Prathapa, Thanveer Ahamed, Janindu Athukorala, Thisara Perera
*Hii.Health, Singapore*

***Abstract***—Hydration influences physiology and daily function, yet repeated practical measurement in everyday settings remains limited. We present initial real-world evidence that a facial-video lifestyle digital biomarker score can track the direction of hydration state transitions. Using Ramadan fasting as a model of prolonged fluid abstinence, we assessed whether the score tracks the transition between the fasting (pre-drink) state and the rehydrated (post-drink) state across three validation layers. In the primary cohort (104 participants, 208 videos ), the score increased from a mean of 36.6 in the fasting (pre-drink) state to 70.6 in the rehydrated (post-drink) state ($\Delta = +33.9$, 95% CI [+30.7, +37.1], $p < 0.001$, Cohen's $d = 2.03$), with 102 of 104 participants (98.1%) showing an upward shift. In a nested body-weight subgroup (20 participants), body weight increased by +0.74 kg after fluid intake ($p < 0.001$), confirming fluid uptake and providing supportive physical anchoring. In a nested serum osmolality subgroup (10 participants), osmolality decreased by −5.1 mOsm/kg ($p < 0.001$), with all post-drink values within the normal clinical range, providing exploratory biological support. An exploratory secondary comparison of water versus electrolyte rehydration showed no statistically significant difference in short-term score change ($p = 0.247$). This study does not establish a clinical diagnostic test for dehydration. Rather, these findings support the score as a lifestyle digital biomarker for tracking the direction of hydration state change in real-world settings.

## I. INTRODUCTION

### *A. The Problem: Why Repeated Hydration Measurement Is Difficult*

Hydration influences cognitive performance, physical endurance, thermoregulation, and renal function [1], yet it remains one of the most difficult physiological states to measure repeatedly in everyday life. Hydration assessment is complex because different markers reflect different compartments, time windows, and physiological assumptions; therefore, no single marker is sufficient across all contexts [2,18]. Because hydration biomarkers vary by context and time course, interpretation often requires marker-specific reference values rather than a single universal threshold [18,19]. Current methods each present practical barriers:

**Blood-based tests** (serum or plasma osmolality) provide the most direct physiological measure but require venipuncture and laboratory analysis, making them impractical outside clinical settings and poorly suited to large-scale repeated assessment [2, 20].

**Urine-based tests** are non-invasive but inconvenient, delayed relative to current hydration state, and subject to dietary and timing confounders [3].

**Body weight** provides an indirect proxy for acute fluid change but cannot distinguish water loss from other mass changes and requires a known baseline [4].

**Thirst perception** is subjective, delayed relative to physiological need, and influenced by age, habit, and environment [5].

**Bioimpedance devices** require specialized hardware, are sensitive to electrode placement, and are poorly suited to frequent repeated use [6].

Furthermore, hydration status is inherently dynamic - it changes continuously with fluid intake, activity, temperature, and metabolic demand. A practical tool must therefore support repeated, low-burden measurement, which none of the above methods achieve in everyday settings.

### *B. The Opportunity*

A practical, repeatable hydration measure could move hydration guidance from vague advice toward measurable physiology, and support research linking hydration patterns to functional outcomes such as cognitive performance, recovery, and fatigue.

### *C. The Solution*

We introduce a lifestyle digital biomarker score derived from a short facial video recording. The score is based on physiological signal patterns associated with remote photoplethysmography (rPPG) - the detection of subtle cardiovascularrelated signals from facial skin color changes captured by a standard camera [7,8,16]. The score is non-invasive, repeatable, and deployable via smartphone without specialized hardware [17].

### *D. Purpose*

Here we evaluate whether the lifestyle digital biomarker score shows directional responsiveness to a real-world hydration state transition as **a hydration state-tracking tool**, rather than validating it as a clinical diagnostic test for dehydration. The validation uses a real-world Ramadan fasting model to assess whether the score tracks the transition between the fasting (pre-drink) state and the rehydrated (post-drink) state, with nested subgroup support from concurrent body-weight change and exploratory serum osmolality change.

## II. OVERVIEW OF THE DIGITAL HYDRATION BIOMARKER SCORE

### *A. What the Score Is*

The lifestyle digital biomarker score is a bounded value generated from a 30-second facial video, reflecting physiological signal patterns associated with hydration state. The score ranges from 0 to 100, where higher values indicate a state more consistent with recent fluid intake. A value of 50 serves as the operational classification boundary - the point at which the model's classification uncertainty is maximal. This boundary is not a clinical cut-off for dehydration; it is an operational threshold for distinguishing the fasting (pre-drink) state from the rehydrated (post-drink) state within this study's context.

### *B. How It Works*

The score is derived from rPPG-associated signal extraction and analyzed using a proprietary AI model built on a deep-learning framework (DeepPhys-1D, 106,082 parameters) [9]. Signals extracted from facial skin regions in the video are processed by the model to identify physiological signal patterns that have been observed to differ between fasting (pre-drink) and rehydrated (post-drink) states. These patterns are associated with cardiovascular and perfusion-related features captured through subtle skin color changes, though the precise physiological mechanisms underlying these differences remain an area of active investigation. The model learns to distinguish these states from these extracted signals without requiring explicit physiological feature engineering.

### *C. Development and Model Lock*

The model was developed using a proprietary labelled dataset of facial video recordings collected across multiple sites in Singapore and Sri Lanka, comprising a substantial number of labelled video samples with representation across different ages, sexes, skin tones, and body compositions given the importance of demographic and skin-tone diversity for evaluating camera-based physiological sensing models [22]. Development involved iterative experimentation across multiple architectures and training configurations, with the final model selected for balanced accuracy and training stability (validation accuracy 90.48%, AUC 0.91 on a held-out evaluation set).

The Ramadan validation cohort described in this paper was completely separate from the model development data and was not used for model development, hyperparameter tuning, threshold selection, or model selection. The score was locked before the validation study began. All scores were generated blind to the before/after condition labels - the model received only the signals extracted from each video, with no metadata about timing or study condition.

## III. REAL-WORLD VALIDATION DESIGN

### *A. Study Model*

Ramadan fasting provides a real-world model of prolonged fluid abstinence. During Ramadan, Muslim participants abstain from all food and water from dawn to sunset - typically 12 to 13 or more hours - generating a naturally occurring fasting (pre-drink) state followed by a rehydration event [10].

### *B. Primary Question*

*Can the lifestyle digital biomarker score track the transition between the fasting (pre-drink) state and the rehydrated (post-drink) state in a real-world setting?*

### *C. Validation Structure*

The study employed a layered design:

**Primary cohort:** 104 participants, 208 facial videos.

**Nested body-weight subgroup:** 20 participants with concurrent weight measurements.

**Nested serum osmolality subgroup:** 10 participants with concurrent venous blood sampling.

Not all participants underwent all measurements because blood testing is invasive and required separate consent. Participants who contributed to the weight and osmolality subgroups were drawn from the same study population, and a subset of participants contributed data to more than one validation layer. Individual-level cross-modality analysis linking digital score change, body-weight change, and serum osmolality change was available only for a small cross-matched subset of 5 participants and is therefore reported separately as exploratory individual-level concordance, in part to preserve participant anonymity and in part because not all subgroup members had complete data across all three modalities. This is a recognized limitation. The validation evidence is therefore presented as directional concordance across tiers - each tier independently shows the expected direction of change - rather than as individual-level correlation.

### *D. Study Format*

This was a pragmatic two-session real-world study. Key protocol parameters are summarized in Table I. Full procedural details are provided in Methods.

**TABLE I**
Study Protocol Summary

| Parameter | Detail |
|---|---|
| **Participants** | 104 adults (≥ 18 years), voluntary recruitment, Ramadan 2026 |
| **Fasting duration** | 12–13+ hours (dawn to sunset) |
| **Fluid volume** | 500 mL per session |
| **Drink conditions** | Session 1: plain water; Session 2: isotonic electrolyte drink |
| **Electrolyte composition** | Sodium: 21 mEq/L; Potassium: 5 mEq/L; Chloride: 16 mEq/L; Citrate: 10 mEq/L |
| **Post-drink scan timing** | ~45 minutes (minor variation due to on-site logistics) |
| **Food before second scan** | Some participants consumed food at Iftar; not controlled |
| **Scan setting** | Majority on-site (standardized device, indoor lighting); subset remote |
| **Posture** | Standing; minor variation at some locations |
| **Video duration** | 30 seconds minimum |

## IV. THE SCORE TRACKS A REAL-WORLD HYDRATION STATE TRANSITION

### *A. Primary Finding*

The lifestyle digital biomarker score showed a substantial paired increase from the fasting (pre-drink) state to the rehydrated (post-drink) state across the full cohort. Table II summarizes the results.

**TABLE II**
Primary Cohort Summary (N = 104)

| | N | Mean | SD | Median | 95% CI |
|---|---|---|---|---|---|
| **Fasting (pre-drink)** | 104 | 36.6 | 11.0 | 38.2 | - |
| **Rehydrated (post-drink)** | 104 | 70.6 | 13.3 | 72.7 | - |
| **Change (Δ)** | 104 | +33.9 | 16.7 | +34.9 | [+30.7, +37.1] |

*Paired t-test: t = 20.66, p < 0.001. Cohen's d = 2.03. Wilcoxon signed-rank test: p < 0.001 (confirming robustness to distributional assumptions).*

A positive score change was observed in 102 of 104 participants (98.1%), indicating that the effect was consistent at the individual level and not driven by group averaging alone. Two participants showed a negative or negligible change, consistent with known individual variability in rehydration timing and fasting-induced physiological response. The mean improvement of +33.9 points represents a large effect (Cohen's d = 2.03).

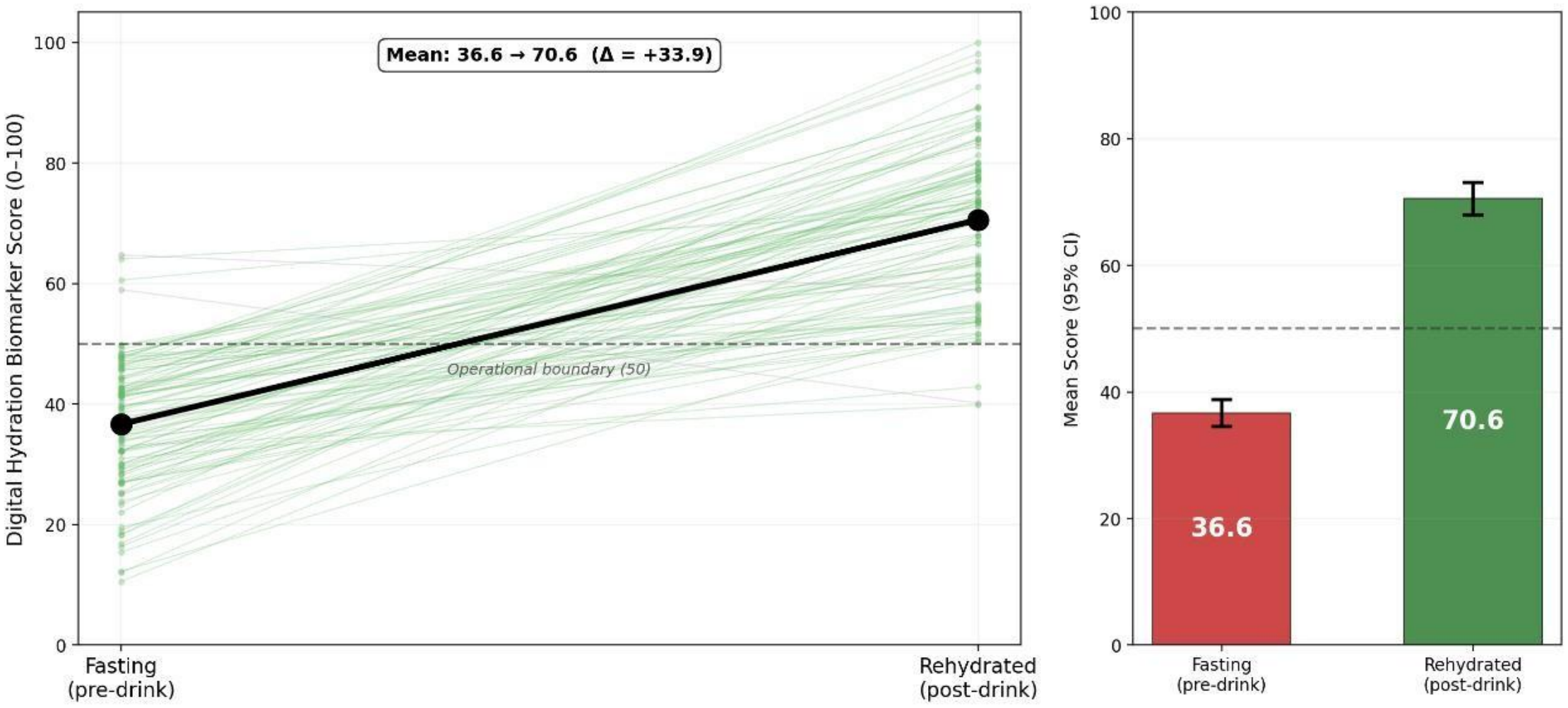

**Fig. 1.** Primary result. Left: individual trajectories (N = 104; green = positive change, grey = negative; black = group mean). Right: group means with 95% CIs. Dashed line = 50-point operational classification boundary.

## V. BODY-WEIGHT SUBGROUP: SUPPORTIVE PHYSICAL ANCHORING

In a nested subgroup of 20 participants, body weight was measured before and after fluid intake. This confirmed that fluid uptake occurred and provided supportive physical anchoring for the primary finding.

**TABLE III**
Body-Weight Subgroup (N = 20)

| | Mean | SD | Range | 95% CI |
|---|---|---|---|---|
| **Weight before (kg)** | 75.3 | 13.0 | 56.4–101.0 | - |
| **Weight after (kg)** | 76.0 | 13.1 | 57.0–102.0 | - |
| **Change (kg)** | +0.74 | 0.20 | +0.4 to +1.1 | [+0.65, +0.83] |

*Paired t-test: t = 16.25, p < 0.001. All 20 participants gained weight (mean +0.98% body weight).*

Every participant gained weight after fluid intake, consistent with acute fluid uptake. Because some participants also consumed food at Iftar before the post-drink measurement, the magnitude of weight increase may reflect both fluid and food intake; however, the consistent positive weight change across all participants still confirms post-intake mass gain and provides supportive physical anchoring for the primary finding. Individual-level linking between weight change and digital score change was not available for all subgroup members, which is a limitation; the concordance reported here is at the subgroup level.

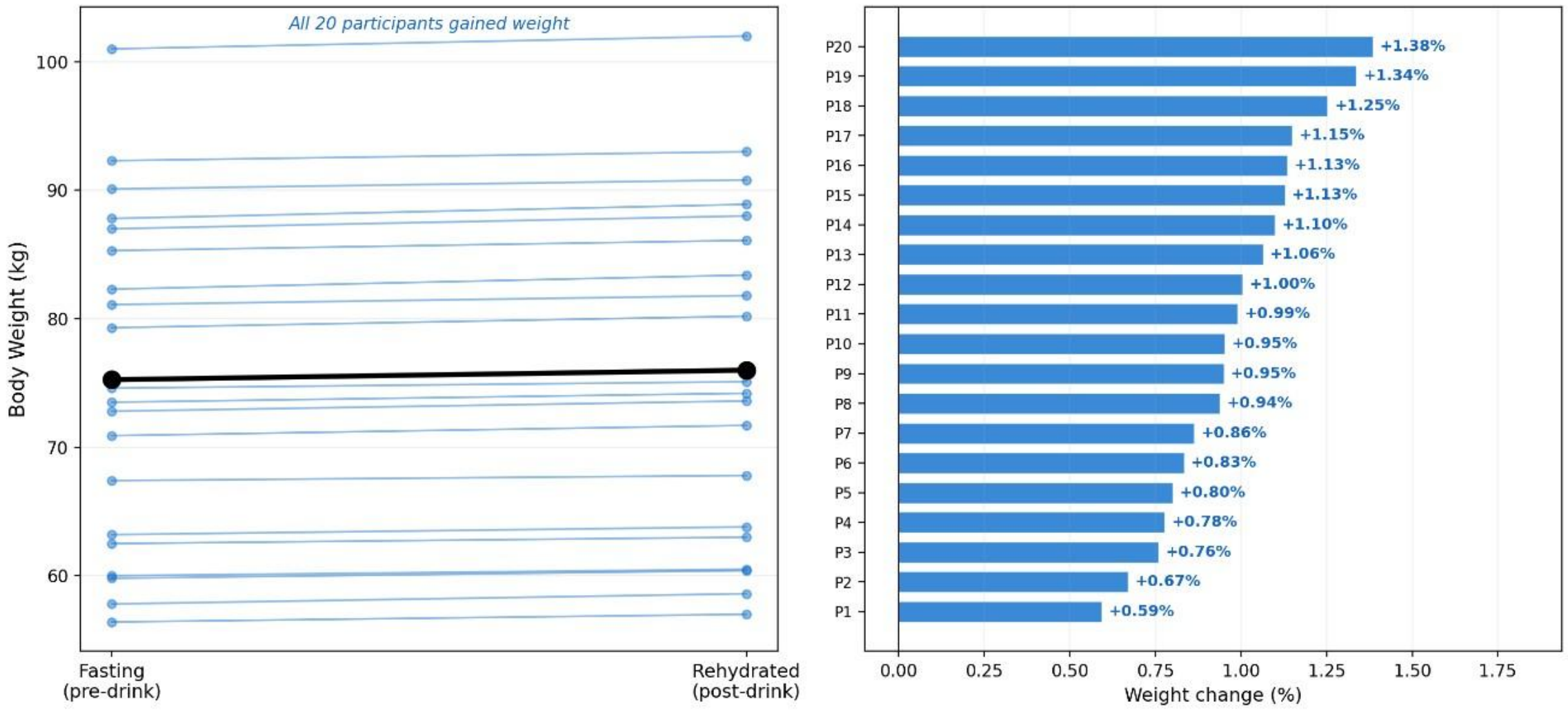

**Fig. 2.** Body-weight subgroup. Left: individual weight trajectories (all positive; black = group mean). Right: individual percentage weight changes, showing consistent fluid uptake across all 20 participants.

## VI. SERUM OSMOLALITY SUBGROUP: EXPLORATORY BIOLOGICAL SUPPORT

In a nested subgroup of 10 participants, venous blood was drawn before and after breaking the fast and serum osmolality was measured by a certified clinical laboratory.

**TABLE IV**
Serum Osmolality Subgroup (N = 10)

| | Mean | SD | Range | 95% CI |
|---|---|---|---|---|
| **Before (mOsm/kg)** | 290.3 | 6.1 | 285–303 | - |
| **After (mOsm/kg)** | 285.1 | 5.1 | 279–295 | - |
| **Change (mOsm/kg)** | −5.1 | 1.7 | −8 to −3 | [−6.2, −4.1] |

*Paired t-test: $t = -9.77$, $p < 0.001$. All 10 decreased. All post-drink values within normal range (275–295 mOsm/kg) [11].*

Serum osmolality decreased in all 10 participants, which is the expected physiological response - fluid intake dilutes blood solute concentration [11]. This decrease occurred in the same study context as the increase in the digital biomarker score, providing exploratory biological support through directional agreement. Individual-level linking was not available for all subgroup members.

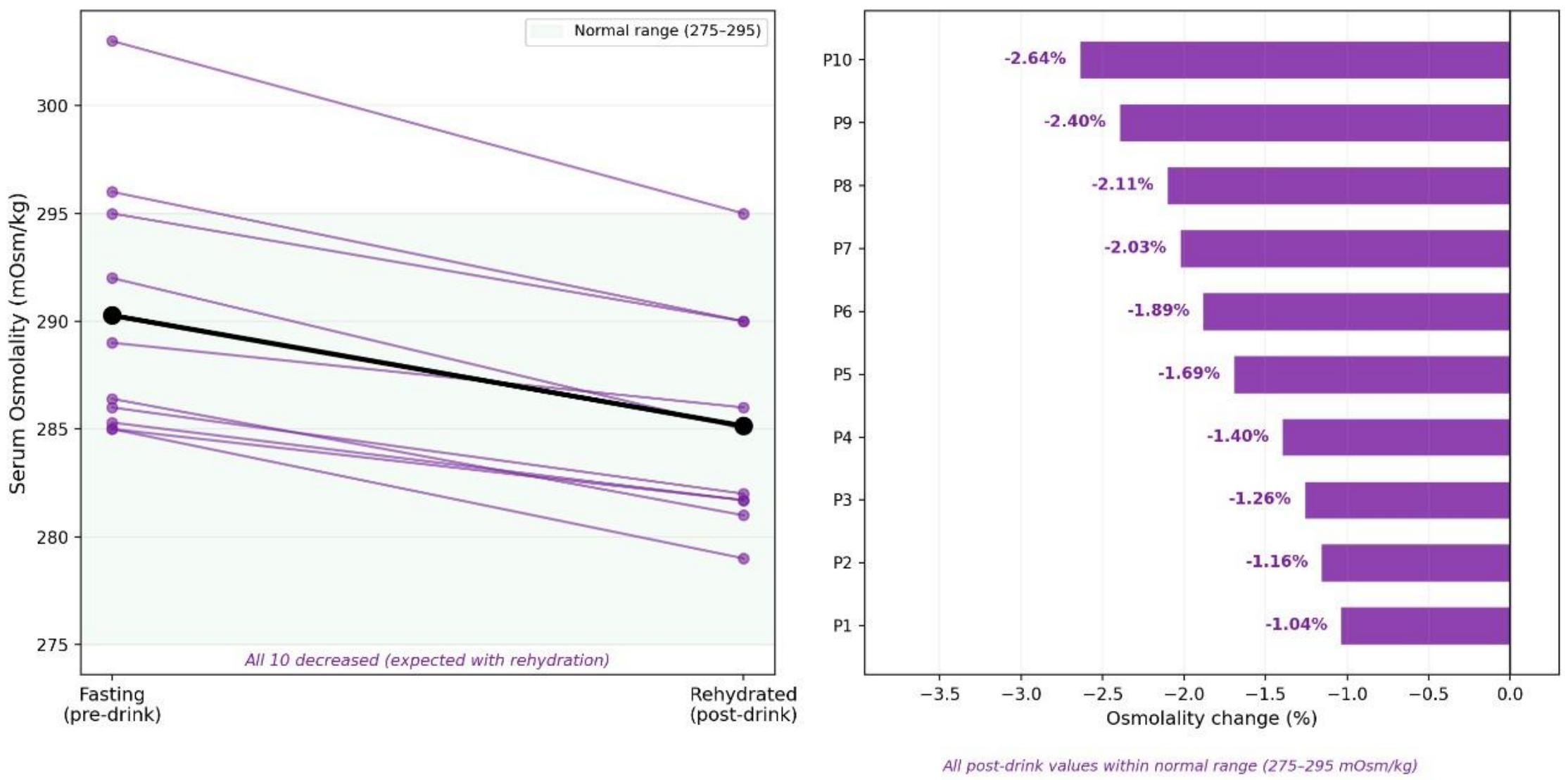

**Fig. 3.** Serum osmolality subgroup. Left: individual trajectories (all decreased; shaded = normal range). Right: individual percentage osmolality changes sorted by magnitude.

## VII. CROSS-MATCHED SUBGROUP: INDIVIDUAL-LEVEL CONCORDANCE

In a small subset of 5 participants, it was possible to link digital biomarker score, body weight, and serum osmolality measurements at the individual level across the fasting (pre-drink) and rehydrated (post-drink) states. This cross-matched subgroup was limited to 5 participants due to the logistical constraints of a pragmatic real-world validation study - coordinating simultaneous video capture, body-weight measurement, and venous blood sampling for the same individual at both time points required substantial effort and was not feasible for the full cohort. Table V presents the cross-matched data.

**TABLE V**

Cross-Matched Subgroup: Individual-Level Data Across Three Modalities (N = 5)

| Subject | Score Before | Score After | Weight Δ (kg) | Weight Δ (%) | Osm Δ (mOsm/kg) | Osm Δ (%) |
|---|---|---|---|---|---|---|
| **A** | 31.0 | 42.9 | +0.50 | +0.8% | −3.3 | −1.2% |
| **B *** | 64.1 | 72.0 | +1.00 | +1.1% | −3.6 | −1.3% |
| **C** | 47.2 | 53.5 | +1.00 | +1.0% | −5.4 | −1.9% |
| **D *** | 60.6 | 74.0 | +1.10 | +1.3% | −8.0 | −2.6% |
| **E** | 45.6 | 61.3 | +0.60 | +0.9% | −7.0 | −2.4% |

*All 5 participants show concordant change across all three modalities: score increased, weight increased, osmolality decreased. * Subjects B and D had fasting (pre-drink) scores above the 50-point operational classification boundary (see Discussion).*

In all 5 cross-matched participants, the digital biomarker score increased, body weight increased, and serum osmolality decreased between the fasting (pre-drink) and rehydrated (post-drink) states. Directional concordance was observed across all three modalities in all 5 cross-matched participants. Fig. 4 shows the individual changes across all three modalities.

**Subjects B and D: fasting (pre-drink) scores above the operational boundary.** Subjects B and D had fasting scores of 64.1 and 60.6 respectively - both above the 50-point operational classification boundary. Despite this, both showed directionally correct score increases after fluid intake (+7.9 and +13.4 points), and both showed concurrent weight gain and osmolality decreases. Subject D is particularly notable: their fasting osmolality was 303 mOsm/kg, above the normal range of 275–295 mOsm/kg, indicating elevated serum osmolality in the fasting state; however, their digital score was 60.6 - above the operational boundary. After drinking, osmolality fell to 295 (within normal range) and the score rose to 74.0.

This suggests that the score's directional sensitivity may be more informative than its absolute classification in some individuals, and that the operational boundary of 50 does not capture every individual's fasting-state physiology. These cases are discussed further in the Discussion.

The availability of cross-matched data for only 5 participants is a limitation, reflecting the logistical constraints of coordinating three concurrent measurement modalities in a self-conducted real-world study. Despite the small sample, the directional concordance across all three modalities in all 5 individuals provides the strongest individual-level evidence in this study that the digital biomarker score tracks a real physiological hydration state change.

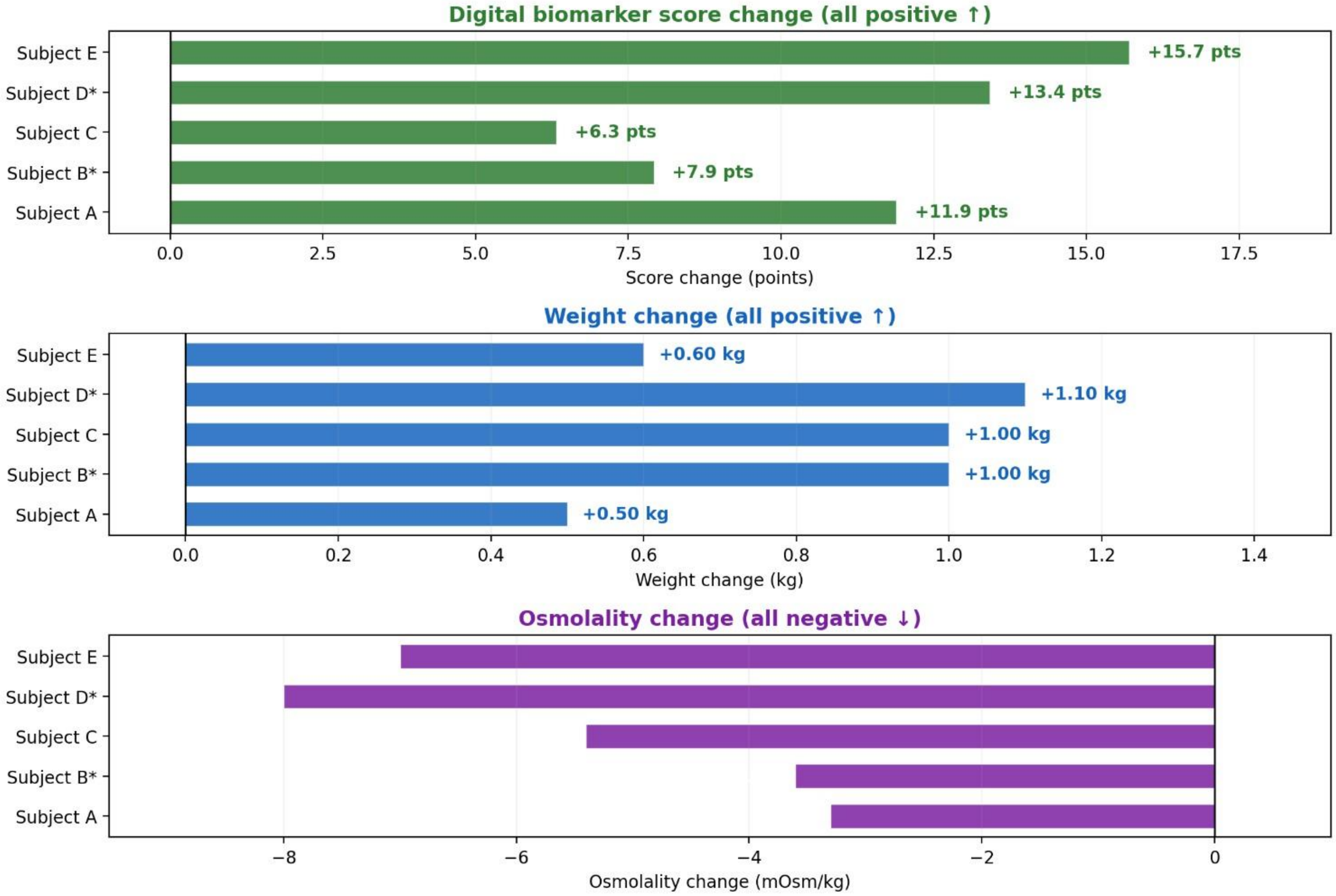

**Fig. 4.** Cross-matched subgroup (N = 5): individual changes across three modalities between the fasting (pre-drink) and rehydrated (post-drink) states. Top: digital biomarker score change (all positive). Middle: body weight change (all positive). Bottom: serum osmolality change (all negative). Each subplot uses its own scale. Asterisks (*) denote Subjects B and D, whose fasting (pre-drink) scores were above the 50-point operational classification boundary.

## VIII. EXPLORATORY SECONDARY COMPARISON: WATER VS ELECTROLYTE

As an exploratory secondary analysis, we compared score improvement between participants who consumed water (N = 50) and electrolyte (N = 54). A subset of participants has contributed to both sessions of water and electrolyte; however, because reliable cross-session pairing of anonymized identifiers was not available, this analysis was treated as exploratory and interpreted at the group level only.

TABLE VI
Exploratory Drink Comparison

| Drink | N | Mean Δ | SD | 95% CI |
|---|---|---|---|---|
| **Water** | 50 | +31.9 | 16.0 | [+27.4, +36.5] |
| **Electrolyte** | 54 | +35.8 | 17.3 | [+31.1, +40.5] |

*Independent t-test: t = −1.17, p = 0.247. Exploratory group-level comparison; statistical independence not fully guaranteed.*

Both water and electrolyte drinks produced substantial short-term score increases, with no statistically significant difference observed in this exploratory group-level comparison. This is consistent with the known physiology of acute rehydration: at approximately 45 minutes post-drink, both fluids have accomplished the primary task of refilling blood plasma volume, while any electrolyte-specific retention advantage - mediated by sodium-dependent renal reabsorption - typically requires 60–120 minutes to manifest [12]. Because some participants may have contributed to both groups without reliable pairing, this comparison should be interpreted with caution.

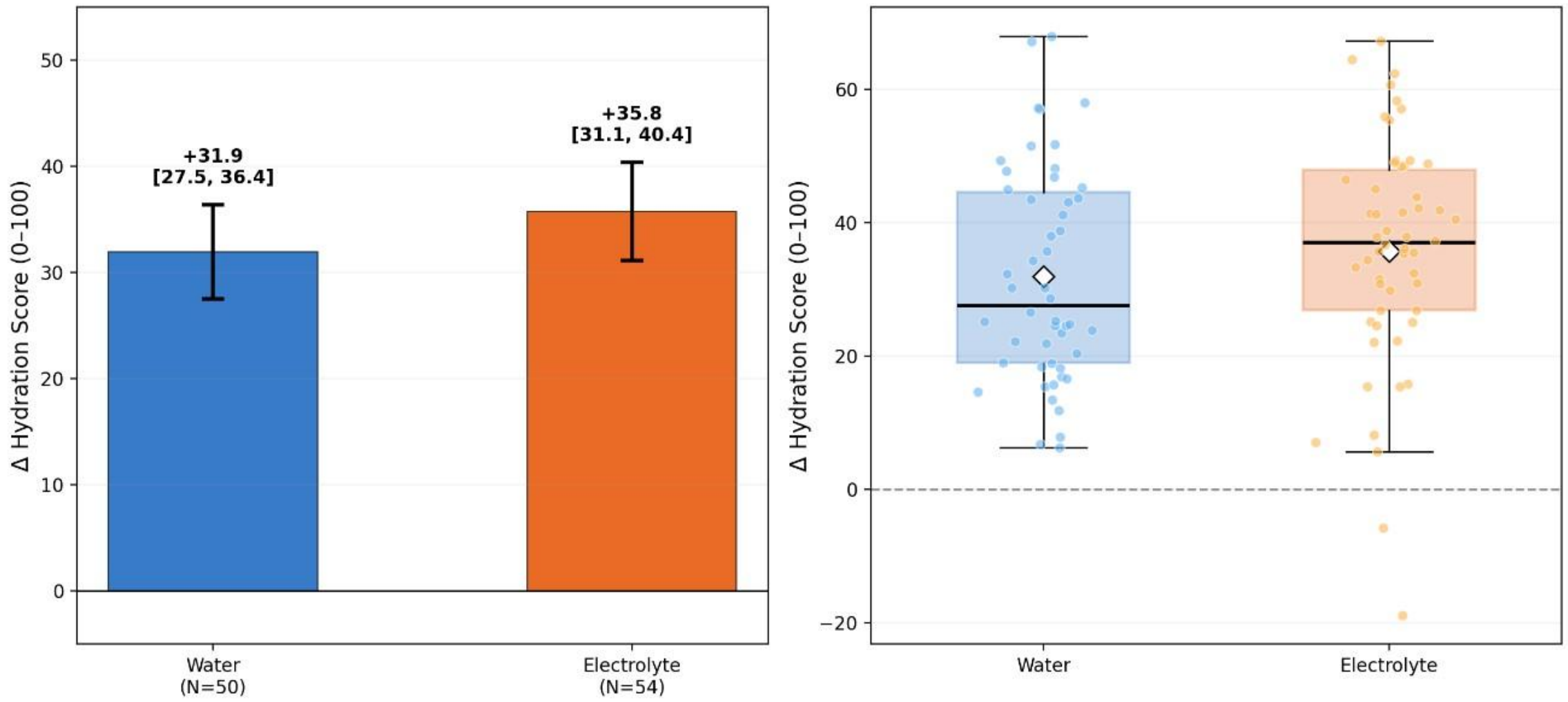

**Fig. 5.** Exploratory drink comparison. Left: mean improvement with 95% CIs. Right: individual distributions. Both conditions show substantial upward shifts.

## IX. SENSITIVITY ANALYSES

To assess the robustness of the primary finding, we conducted several pre-planned sensitivity analyses.

### *A. Non-Parametric Confirmation*

A Wilcoxon signed-rank test confirmed the primary paired result ($p < 0.001$), demonstrating that the finding is not dependent on normality assumptions.

### *B. Drink-Stratified Analyses*

When restricted to water-only participants (N = 50), the mean score change was +31.9 (95% CI [+27.4, +36.5], $p < 0.001$). When restricted to electrolyte-only participants (N = 54), the mean change was +35.8 (95% CI [+31.1, +40.5], $p < 0.001$). The primary finding holds independently within each drink condition.

### *C. Classification Boundary Sensitivity*

The primary analysis uses 50 as the operational classification boundary. To assess sensitivity, we examined the proportion of participants whose fasting (pre-drink) score fell below, and rehydrated (post-drink) score exceeded, alternative boundaries of 40 and 60. At a boundary of 40, the large majority of participants transitioned from below to above. At a boundary of 60, a substantial proportion still showed the transition, though with reduced classification separation. The primary result is not critically dependent on the choice of 50 as the boundary.

#### *D. Distribution of Individual Changes*

The distribution of individual score changes was approximately symmetric with a median of +34.9 and interquartile range of [+23.2, +46.8]. The two participants with negative changes had small magnitudes, and their inclusion does not meaningfully affect the group-level result.

## X. DISCUSSION

#### *A. Principal Finding*

The lifestyle digital biomarker score showed a substantial paired increase from the fasting (pre-drink) state to the rehydrated (post-drink) state in a real-world Ramadan fasting model, with a large effect size (Cohen's d = 2.03) and 102 of 104 participants showing a positive change. This is the core finding: the score tracks the expected direction of a real-world hydration state transition.

#### *B. Lifestyle State-Tracking, Not Clinical Diagnosis*

This study validates the score as a tool for tracking the direction of hydration state change in everyday settings. It does not establish a clinical diagnostic test for dehydration, nor does it claim that the digital score and serum osmolality are interchangeable. The distinction is important: a lifestyle tool needs to reliably detect the direction of change; a clinical diagnostic would require absolute accuracy against a gold standard.

#### *C. The Main Evidence Is Paired Score Movement*

The primary paired analysis (N = 104) is the strongest evidence. Each participant served as their own control, eliminating individual baseline differences. The within-subject design is the appropriate test for a state-tracking biomarker [13].

#### *D. Weight and Osmolality Are Supportive, Not Definitive*

The weight and osmolality subgroups provide directional concordance in the same study context. They are supportive and exploratory, not definitive standalone validations. For the majority of participants, individual-level cross-modality data were not available due to logistical constraints of coordinating simultaneous video, weight, and blood measurements. This limits the strength of within-subject concordance claims across the full cohort. Future studies should be designed to collect concurrent measurements across all modalities from the same individuals.

#### *E. Cross-Matched Subgroup: Individual-Level Evidence*

The cross-matched subgroup of 5 participants provides the strongest individual-level evidence in this study. All 5 showed concordant change across all three modalities: the digital biomarker score increased, body weight increased, and serum osmolality decreased. This individual-level directional concordance partially addresses an important limitation of the tier-level validation design and supports the interpretation that the score is tracking the same broad physiological transition reflected by weight and osmolality changes.

Subjects B and D merit specific discussion. Both had fasting (pre-drink) scores above the 50-point operational classification boundary (64.1 and 60.6 respectively), meaning the model classified them as being in a rehydrated state even before drinking. However, their concurrent physiological measurements tell a different story: both gained weight after drinking (+1.0 kg and +1.1 kg), and both showed osmolality decreases (−3.6 and −8.0 mOsm/kg). Subject D's fasting osmolality of 303 mOsm/kg was above the normal clinical range, indicating elevated serum osmolality in the fasting state - yet the digital score classified this individual as rehydrated. After drinking, Subject D's osmolality fell to 295 (within normal range) and the score rose further to 74.0.

These cases illustrate two important points. First, the score's directional sensitivity - its ability to track the direction of change - appears to be more robust than its absolute classification accuracy at the individual level. Both B and D showed clear upward score movement that was concordant with weight and osmolality changes, even though their absolute fasting scores were above the operational boundary. Second, the fixed boundary of 50 does not capture every individual's fastingstate physiology. Some individuals may maintain baseline physiological patterns that the model interprets as hydrationconsistent even during prolonged fasting - potentially reflecting individual variation in vascular tone, habitual hydration status, or pre-fast fluid loading. Future work should explore whether individualized baselines or adaptive thresholds could improve classification accuracy for such individuals.

The availability of cross-matched data for only 5 of 104 participants is a clear limitation, reflecting the logistical constraints of a pragmatic real-world validation study where data collection, participant coordination, and biological sample handling were managed by the study team. Expanding this cross-matched subgroup is a high priority for future work.

### *F. Real-World Design: Strength and Limitation*

The pragmatic format improves ecological validity - the score was tested in the conditions it is designed for - but introduces variability. Food intake was not controlled, post-drink timing varied slightly, and a subset of data was collected remotely. In particular, some participants consumed food alongside fluid at Iftar before the post-drink scan. This means the observed score change reflects the real-world transition from prolonged fasting to the post-Iftar rehydrated state, not the isolated physiological effect of water intake alone. We consider this appropriate for a lifestyle biomarker validation: in real-world use, people typically eat and drink together when breaking a fast, and the tool must function in that context. Food intake is reported as an uncontrolled variable and should be standardized in future controlled studies. Despite this variability, the score increased in 102 of 104 participants with a large effect size, supported by concurrent weight and osmolality changes, demonstrating that the signal is robust under real-world conditions [14].

### *G. Future Work*

Future work should prioritize: fixed post-drink timing; concurrent individual-level measurement across all modalities; larger blood-sampling subsets; extended observation windows (60–120 minutes); tighter meal standardization; broader demographic testing; and linkage to functional outcomes. A structured daily monitoring protocol with repeated measurements across a full waking day is under development.

### *H. Non-Claims*

This paper does not claim clinical diagnostic accuracy, interchangeability with osmolality, or a definitive drink comparison. The water-versus-electrolyte analysis is exploratory only.

## XI. CONCLUSION

The lifestyle digital biomarker score showed a substantial paired increase from the fasting (pre-drink) state to the rehydrated (post-drink) state in 104 participants (102/104 positive, Cohen's d = 2.03). Body-weight increase (N = 20) confirmed fluid intake. Serum osmolality decrease (N = 10) provided exploratory biological support. In a cross-matched subgroup of 5 participants with all three measurements, directional concordance was observed in every case: the score increased, weight increased, and osmolality decreased. Both water and electrolyte drinks produced substantial short-term score increases, with no statistically significant difference observed in this exploratory comparison. These findings support the score as a lifestyle digital biomarker for tracking hydration state change in real-world settings.

## XII. METHODS

### *A. Score Generation*

Each participant recorded a 30-second facial video using a smartphone camera under normal resting conditions. Facial region of interest (ROI) was extracted using MediaPipe face detection and landmark estimation [15]. The rPPG-associated physiological signal was extracted from the ROI and processed by a locked proprietary AI model built on the DeepPhys1D deep-learning framework (106,082 parameters) [9] to produce a hydration probability, linearly scaled to the 0–100 score. The model was developed using a proprietary labelled dataset collected across sites in Singapore and Sri Lanka and was locked before this validation began. The analysis sequence was strictly ordered: model locked first, then validation scores generated from the recorded videos, then statistical analysis performed. No model parameters, thresholds, or architectural choices were modified after observing any validation results. All scores were generated blind to the before/after condition labels.

### B. Participants

Adult Muslim volunteers (age ≥ 18, range approximately 18–60 years) observing Ramadan fasting in 2026 were recruited through voluntary sign-up. Participants received meals and refreshments after breaking their fast. Exclusion criteria: age below 18, unwillingness to provide consent, or inability to record a video meeting quality requirement. Participants in the blood-sampling subgroup provided separate consent for venipuncture.

### C. Protocol

A 30-second facial video was recorded immediately before breaking the fast. Participants then consumed 500 mL of the assigned study fluid - plain water (Session 1) or isotonic electrolyte drink (Session 2; Sodium: 21 mEq/L; Potassium: 5 mEq/L; Chloride: 16 mEq/L; Citrate: 10 mEq/L). A second video was recorded approximately 45 minutes later. Some participants consumed food alongside fluid at Iftar; this was not controlled and is reported as a limitation. On-site recordings used a single standardized device under controlled indoor lighting, with participants in a standing position under normal resting conditions. Remote participants followed a standardized remote capture protocol: they were provided step-by-step instructions covering device setup (consistent device model where possible), lighting conditions (consistent indoor lighting, avoiding direct sunlight or backlighting), facial positioning (centered, still, neutral expression), and posture (standing). In the body-weight subgroup, participants were weighed on a calibrated scale before and after fluid intake. In the serum osmolality subgroup, venous blood was drawn at both time points by a trained phlebotomist and analyzed by freezing point osmometry at a certified clinical laboratory.

### D. Quality Control and Data Exclusion

Videos were subjected to quality gating before analysis. Minimum duration was 30 seconds. Frames with excessive motion, lighting artefacts, or facial occlusion were excluded using automated frame-level quality checks. Participants with incomplete or technically inadequate recordings were excluded prior to scoring. Only complete before/after pairs were included in the analysis; no imputation was performed and no data were selectively excluded based on score values. Subgroup analyses (weight, osmolality) used only participants with valid paired subgroup data. These quality-control and data-handling procedures were applied uniformly and were not modified after observing any results.

### E. Statistical Analysis

Paired t-tests assessed within-subject fasting (pre-drink) versus rehydrated (post-drink) differences. A Wilcoxon signed-rank test was performed as a non-parametric sensitivity check. Cohen's d was computed as the primary effect-size measure. Confidence intervals were computed using the t-distribution for the relevant sample size. An independent-samples t-test was used for the exploratory drink comparison. Analyses were performed in Python (scipy.stats). No corrections for multiple comparisons were applied as each analysis addresses a distinct pre-specified question.

### F. Ethics

All participation was voluntary with informed consent. Blood-sampling participants provided separate consent. All data were anonymized prior to analysis.